\documentclass[a4paper]{article}

\usepackage{INTERSPEECH2016}
\usepackage{subcaption}
\usepackage{graphicx}
\usepackage{amssymb,amsmath,bm}
\usepackage{textcomp}
\usepackage{multirow}

\ninept

\title{LANDMARK-BASED CONSONANT VOICING DETECTION ON MULTILINGUAL CORPORA}

\makeatletter
\def\name#1{\gdef\@name{#1\\}}
\makeatother \name{{\em Xiang Kong$^1$, Xuesong Yang$^2$, Jeung-Yoon Choi$^3$,}\\ {\em Mark Hasegawa-Johnson$^2$, Stefanie Shattuck-Hufnagel$^3$}}

\address{$^1$Computer Science Department, University of Illinois at Urbana-Champaign\\
  $^2$Beckman Institute, University of Illinois at Urbana-Champaign\\
  $^3$Speech Communication Group, Research Laboratory of Electronics, MIT\\
  {\small \tt \{xkong12, xyang45, jhasegaw\}@illinois.edu, \{jyechoi, sshuf\}@mit.edu}
}

\begin{document}
  \maketitle
  \begin{abstract}
    This paper tests the hypothesis that distinctive feature classifiers anchored at phonetic landmarks can be transferred cross-lingually without loss of accuracy.  Three consonant voicing classifiers were developed: (1) manually selected acoustic features anchored at a phonetic landmark, (2) MFCCs (either averaged across the segment or anchored at the landmark), and (3) acoustic features computed using a convolutional neural network (CNN).  All detectors are trained on English data (TIMIT), and tested on English, Turkish, and Spanish (performance measured using F1 and accuracy). Experiments demonstrate that manual features outperform all MFCC classifiers, while CNN features outperform both. MFCC-based classifiers suffer an overall error rate increase of up to 96.1\% when generalized from English to other languages. Manual features suffer only an up to 35.2\% relative error rate increase, and CNN features actually perform the best on Turkish and Spanish, demonstrating that features capable of representing long-term spectral dynamics (CNN and landmark-based features) are able to generalize cross-lingually with little or no loss of accuracy.
  \end{abstract}

  \section{Introduction}

In contrast to the conventional data-driven speech recognition model,
acoustic correlates of distinctive features are found in an acoustics
phonetic recognizer~\cite{stevens1} so as to extract interpretable
acoustic information. There are two types of distinctive features in
this model: articulator-free feature and articulator-bound
feature. Articulator-free features determine phone manner class, while
articulator-bound features specify the phone identity.  Since features
in this model depend on the properties of the vocal tract, they are,
to some extent, universal and independent of the language being
spoken. For obstruent consonants in English, e.g. fricatives,
affricates, and stops, three articulators ([lips], [tongue body] and
[tongue blade]) can form the constriction to produce
consonants. Obstruent consonants are further categorized by consonant
voicing which can be described by the articulator-bound feature [stiff
  vocal folds]~\cite{stevens2}.
    
Much related work has been done about consonant acoustic and
voicing. Shadle~\cite{Shadle} studied fricative consonants using
mechanical models, theoretical models, and acoustic analysis, and
found that the most important parameters for fricatives are the length
of the front cavity, the presence of an obstacle and the flow
rate. Speech production mechanism differences between voice and
voiceless stops are mainly due to muscle activity, which relaxes the
tongue root during voiced stops, altering aerodynamics near the vocal
folds in order to maintain voicing during closure~\cite{bell}. Vowel
cues (i.e. vocalic duration and F1 offset frequency) are also
correlates of consonant voicing \cite{cv}. A module for detecting
consonant voicing based on these acoustic
correlates~\cite{choi1999detection} first determines acoustic
properties according to consonant production, then extracts acoustic cues,
and classifies them to detect consonant voicing.

One of the traditional methods to detect consonant voicing uses mel-frequency cepstral coefficients (MFCCs)~\cite{pa,cooke2008interspeech},
e.g.~voicing can be detected with 74.7\% to 80\% overall accuracy~\cite{rdm,bush}.
Overall, good performance of MFCC-based method on consonant voicing is
possible,however, MFCCs mostly codify the ``filter'' information in the
source-filter theory of speech production, and are therefore less efficient
in capturing information about the voice source.

In contrast, much of the consonant voicing information can be captured
in the characteristics of the vocal fold vibration patterns, therefore
capturing acoustic phonetic features indicative of vocal fold
vibration has the potential to measure consonant voicing.  Though
voicing does not continue uninterrupted during obstruent closure in
English, there are striking differences near consonant closure and
release landmarks.  Landmarks~\cite{stevens1} identify times when the
acoustic patterns of the linguistically motivated distinctive features
are most salient; acoustic cues extracted in the vicinity of landmarks
may therefore be more informative for the classification of
distinctive features than cues extracted from other times in the
signal.  To the best of our knowledge, the highest accuracy for
voicing classification of obstruents uses acoustic features extracted
with reference to phonetic landmarks, with accuracies of 95\% and
96\%~\cite{ali1,ali2} for stops and fricatives respectively.
  
The choice of data representation is essential for the performance of
detection or classification tasks. Discriminative information from raw
data can be extracted by taking advantage of human perceptual
ingenuity and human prior knowledge. However, the process of designing
these manual features is laborious and time-consuming. Deep learning
techniques transform raw data into multiple levels of abstraction by
stacking multiple layers with non-linearities, thus learning complex
features automatically~\cite{lecun2015deep}.  Though the accuracy of
speech recognizers built from deep networks is
high~\cite{hinton2012deep}, results on the cross-language portability
of deep networks include both positive and negative outcomes.  We
propose that deep networks trained to classify distinctive
features should be cross-language portable, because of the
universality of the features they are trained to classify.

In order to test the hypothesis that distinctive features anchored at phonetic landmarks can be transferred cross-lingually, our models that are trained on an English corpus will be applied for the detection tasks of three different languages, including English, Spanish, and Turkish.
    
In this paper, acoustic landmark theory and the definition of its regions are described in Section 2. Within these landmark regions, section 3 illustrates acoustic feature representations that help to improve the performance of voicing detection, consisting of manually designed acoustic cues and features learned from deep neural networks. Experiments are described in Section 4, Results in Section 5, and Conclusions in Section 6.

  \section{ ACOUSTIC LANDMARKS AND DISTINCTIVE FEATURES}

Landmarks \cite{stevens1} are defined as points in an utterance around
which information about the underlying distinctive features may be
extracted. Four types of landmarks were proposed in \cite{stevens1}:
Vowel (V), Consonant release (Cr), Consonant closure (Cc), and Glide
(G).  Cr landmarks and Cc landmarks are further specialized by manner
class: S=stop, F=fricative, N=nasal.

In this work, we assume that we have been given the right landmark
positions in a speech signal.  To convert TIMIT phonetic
transcriptions to landmark transcriptions, the following rules were
used. Each stop release segment has one Sr landmark at its start time;
each stop closure segment has one Sc landmark at its start time; and
each affricate, fricative, or nasal has a Cc landmark at its start
time and a Cr landmark at its end time (where C=F for affricates and
fricatives, C=N for nasals).  Vowels and glides each have one
landmark, located at the midpoint.

TIMIT label files specify the start time and the end time of each
phone, from which landmark locations were computed and generate
landmark files. Table \ref{tab:landmark_label} illustrates examples of
acoustic landmarks extracted from TIMIT. The first column denotes
landmark time, the second column landmark type.  The first row in
Table \ref{tab:landmark_label}, for example, shows that a fricative
closure landmark happens at time alignment 0.916s.
   
\begin{table}[htbp]
      \caption{\label{tab:landmark_label} {\it Examples of landmark transcription from TIMIT.}}
      \vspace{-2mm}
      \centerline{
        \begin{tabular}{| c | c |}
          \hline
          \multicolumn{1}{|c|}{Time (s)} & 
          \multicolumn{1}{c|}{Landmark type} \\
          \hline 
           0.1916 & Fc \\
           0.2839 & Fr\\
           0.3213 & V\\
           0.3864 & G\\
          \hline
        \end{tabular}
      }
      \vspace{-2mm}
\end{table}
  
Stevens \cite{stevens1} proposed that distinctive features obtained from closure and release landmark regions should be universal across languages. Motivated by this theory, landmark positions across multiple languages can be labeled by the same rules as above; this paper tests corpora in Spanish and Turkish. After finding landmark positions, we denote the landmark regions as follows~\cite{yang2016landmark}:
  \begin{itemize}
    \item If a Cc landmark is located the start of one phone, speech signals after that time (+20ms) are extracted.
    \vspace{-1mm}
    \item If a Cr landmark is at the start of one phone, speech signals before that time (-20ms) are extracted.
  \end{itemize} 
 
After finding landmarks, acoustic cues which could be used to
determine distinctive features are extracted. Distinctive features are
a concise description of subsegmental attributes of a phone, with
a relatively direct relation to acoustics and articulation. They take
on binary values and form a minimal set which can distinguish each
segment from all others in a language, therefore if we can determine
distinctive features, the phonetic transcription is thereby determined.

\section{ACOUSTIC FEATURE REPRESENTATIONS}

Within each landmark region, the acoustic features are extracted for
discriminating voiced consonants from unvoiced ones, including
manually designed acoustic cues (summary see Table~\ref{tab:acoustic_cues}) and the features learned from deep
neural networks.

\subsection{ Manually Designed Acoustic Cues}
    
\emph{\textbf{Voice onset time (VOT)}}: At first we found duration of
every consonant (e.g. stop, fricative and affricate). Duration
refers to the length between the release and the onset of voicing. For
stop release segments, duration is the voice onset time (VOT)
\cite{rdm} which carries voicing information about English stops.
Duration also carries voicing information for fricatives and
affricates, both of which are shorter if voiced than unvoiced.

\emph{\textbf{Peak normalized value of the cross-correlation (PNCC)}}:
Increasing cross correlation value will exist when producing voiced
consonants (stops, fricatives and affricates) [\cite{stevens1}. PNCC
  is originally denoted in \cite{talkin1995robust}, and besides using
  normalized cross-correlation, we retain its peak, which
  captures cross correlation value transitions. Therefore, we used the
  peak normalized value of the cross correlation (PNCC) as another
  acoustic feature for consonant voicing.

\emph{\textbf{Amplitude of fundamental frequency (H1)}}: Although the
amplitude of the speech signal varies with time, the amplitude of
unvoiced speech utterances is much lower than that of voiced
segments. Therefore, to distinguish the strength of vocal fold
vibration, the amplitude of fundamental frequency (H1) is a reasonable
third acoustic feature for consonant voicing.

\emph{\textbf{Formant transitions}}: Formant
transitions~\cite{stevens1974role} are different for voiced and
unvoiced consonants.  There is a significant formant transition
present following voice onset in voiced obstruents, less so for
unvoiced obstruents; the difference is especially marked for stops,
but is also observable for other obstruents.  Formant transitions are
therefore also retained.

\emph{\textbf{Energy}}: Energy of voiced and unvoiced consonants is
also different.  As voice energy can be observable at both low and
relatively high frequencies, while unvoiced energy is concentrated at
high frequencies, the last four acoustic features are the
root-mean-square (RMS) energy, energy between 0-400Hz (E1), energy
between 2000-7000Hz (E2), and the ratio of E1 and E2.
     
  \begin{table}[th]
    \caption{\label{tab:acoustic_cues} {\it Acoustic features in landmark regions}}
    \vspace{-2mm}
    \centerline{
      \begin{tabular}{| p{4cm} | p{3.8cm} |}
        \hline
        Feature & Brief reason \\
        \hline 
        1. RMS energy & \multirow{4}{4cm}{To assess the differences of energy levels between low and high frequency} \\
        2. 0-400Hz energy (E1) &\\
        3. 2000-7000 Hz energy (E2)&\\
        4. Ratio of E1 and E2&\\
        \hline
        5. Peak normalized cross correlation (PNCC) & When vocal folds vibrate, PNCC increases\\
        \hline
        6. Amplitude of first harmonic  & Access the strength the vocal fold vibration\\
        \hline
        7. Voice onset time (VOT) & \multirow{3}{4cm}{Carries voicing information for stop and duration of fricative; differ for v/u fricative}\\
        8. Formants transition & \\
        &\\
        \hline
      \end{tabular}
    }
  \end{table}

  \subsection {Convolutional neural networks (CNN)}

  A common raw feature representation of speech signals as inputs is the magnitude of log-scale mel filterbanks over time. However, this paper proposes to extract features across a landmark region that is too short (20ms) for multiple frames, therefore 1D discrete Fourier transformation, and 1D filterbanks are considered as inputs. 

\begin{figure}[t]
    \centering
    \includegraphics[scale=0.5]{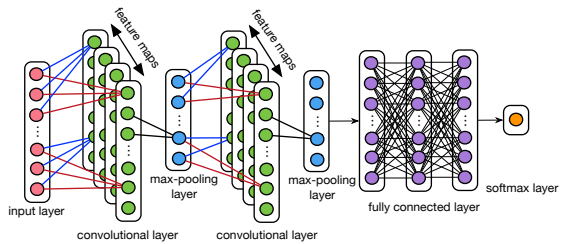}
    \caption{Convolutional neural networks for speech signals.}
    \label{fig:cnn}
  \end{figure}

  Figure \ref{fig:cnn} illustrates the architecture of convolutional neural networks that consist of three types of layers--convolution, max-pooling, and fully connected layers. In a convolutional layer, each neuron takes as inputs local patterns in the previous layer. All neurons in the same feature map share the same weight matrix. A max-pooling layer is stacked following each convolutional layer that similarly takes local patterns as inputs, and down-samples to generate a single output for that local region. Multiple fully connected layers are concatenated after multiple building blocks of convolutional-pooling pairs. A softmax layer with a single neuron is taken as the output that capture the posterior probability of the positive label (voicing). During back-propagation, a first-order gradient-based optimization method based on adaptive estimates of lower-order moments (\emph{Adam}) \cite{kingma2014adam} is used.

  \section{EXPERIMENTS}

  \subsection{Multilingual Corpora}
  Three language corpora, including American English, Spanish, and Turkish, are considered. Only the TIMIT TRAIN corpus is chosen for model training, while others are selected as test sets. Table \ref{tab:corpus} illustrates the number of samples in these corpora.
      
  English Corpus: TIMIT \cite{garofalo1993darpa} corpus contains broadband recordings of 630 speakers of eight major dialects of American English, each reading ten phonetically rich sentences and includes time-aligned orthographic, phonetic and word transcriptions, as well as a 16-bit, 16kHz speech waveform file for each utterance. 
      
  Spanish corpus: The phonetic Albayzin corpus of Spanish is divided into two subcorpora, one for training and another for testing purposes, because it was initially developed to train speech recognition engines. The training subcorpus is made of 200 phrases; 4 speakers produced all 200 phrases and 160 speakers produced 25 out of these 200, so the set of 200 phrases is produced 24 times. The phrases are acoustically balanced, according to a statistical study of the frequency of each sound in Castillian Spanish.
      
  Turkish corpus: Middle East Technical University Turkish Microphone Speech Corpus (METU) \cite{salor2002developing} was selected as Turkish test set.  120 speakers (60 male and 60 female) speak 40 sentences each (approximately 300 words per speaker), which makes around 500 minutes of speech in total. The 40 sentences are selected randomly for each speaker from a triphone-balanced set of 2462 Turkish sentences.
  \begin{table}[th]
    \caption{\label{tab:corpus} {\it Voiced class distribution on multilingual corpora}}
    \vspace{-2mm}
    \centerline{
      \begin{tabular}{| c| c | c |}
        \hline
        &Voiced consonant & Unvoiced consonant   \\
        \hline 
         TIMIT (train)& 56,269& 40,475  \\
         TIMIT (test)&  20,769  &14,214 \\
         Spanish (test)&  68,946& 24,529  \\
         Turkish (test)&  13,179& 4,722 \\  
        \hline
      \end{tabular}
    }
    \vspace{-4mm}
  \end{table}

  \subsection{Feature Extraction}
  The calculation of manual acoustic features anchored at a phonetic landmark region, and MFCCs (either averaged across the phonetic segment or anchored at the landmark region), and raw data representation of speech signals in landmark regions are illustrated as follows, respectively.

  \emph{\textbf{MFCCs}}: A Hamming window is first applied, with duration equal either to the landmark region, or to the duration of the whole phone.  The windowed signal is then transformed to compute MFCC(13) or MFCC(39).
  
  \emph{\textbf{Duration, formant, transition, PNCC and H1}}: a robust RAPT \cite{talkin1995robust} algorithm for pitch tracking that is based on normalized cross-correlation and dynamic programming is applied using Wavesurfer\footnote{\url{http://www.speech.kth.se/wavesurfer/}}. The fundamental frequency, probability of voicing (1.0 means voiced and 0.0 means unvoiced), local error of the pitch, and the peak normalized value of the cross correlation (PNCC) are obtained. After getting the pitch for each landmark segment,  FFT amplitude at the pitch frequency was measured, and FFT spectra were used to measure formant transitions.

  \emph{\textbf{Energy}}: Butterworth filter is used to design a
  bandpass filter with $\le~3$dB of passband ripple and $\ge~40$dB
  attenuation in the stopbands, then energy of filtered signals is
  computed in the bands 0-400Hz and 2000Hz-7000Hz.
  
  \emph{\textbf{Raw features for neural networks}}: 1024 point magnitude FFT is performed. Mel-scale filterbank features are calculated by multiplying frequency response with a set of 40 triangular bandpass filters equally spaced in Mel frequency.
  In order to apply the early stopping strategy during training procedure, a held-out development set (10\%) is stratified sampled from training set. The training will stop when the validation loss is not decreasing anymore within 10 consecutive epochs.

\begin{figure}[htbp]
  \centering
  \includegraphics[width=\linewidth]{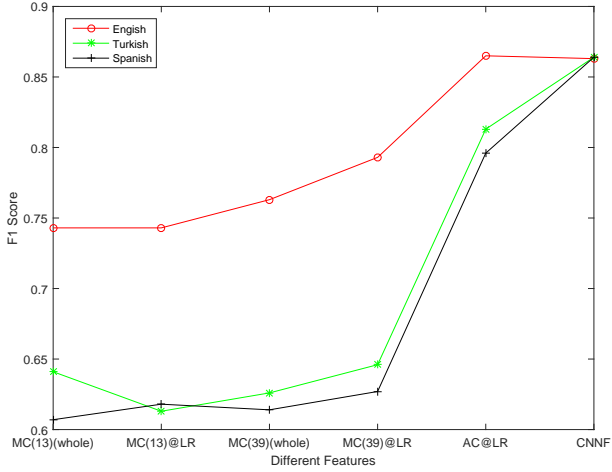}
  \caption{F1 score for three different features (MC:MFCC, AC:acoustic cues, and CNN) on three test languages (English, Spanish, and Turkish).  MFCCs tested at LR:landmark region, whole:whole phone.}
  \label{fig:sfig1}
\end{figure}
\begin{figure}[htbp]
 \centering
 \includegraphics[width=\linewidth]{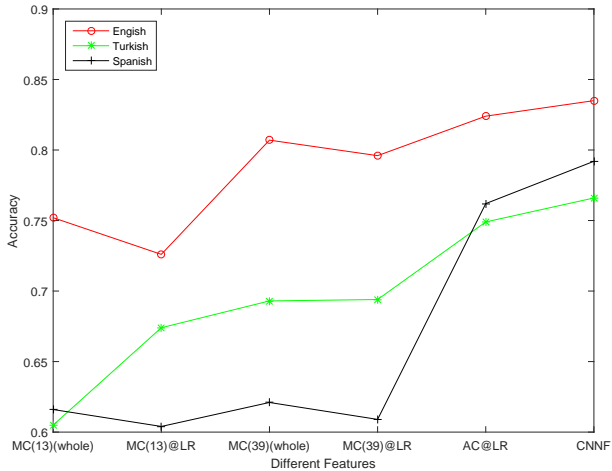}
 \caption{Overall accuracy for MC, AC, and CNN on three test languages.}
 \label{fig:sfig2}
 \end{figure}
\begin{figure}[htbp]
\includegraphics[width=\linewidth]{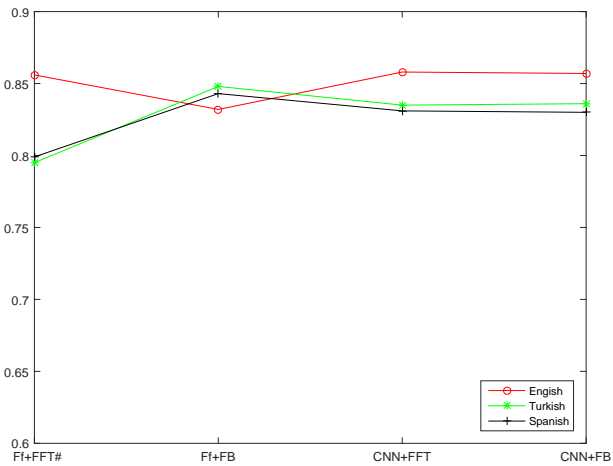}
   \caption{Overall accuracy for different neural network configurations (Ff: Feedforward neural network, FB: Filterbank).}
   \label{fig:sfig3}
\end{figure}

\section{RESULTS}

Consonant voicing is detected using MFCCs, acoustic cues, and CNNs. These models are all trained in English, and tested on English, Spanish, and Turkish, respectively. Support vector machine with radial basis function kernel is used as the binary classifier based on acoustic features, while CNNs are used as an end-to-end classifier. The F1 score of voicing consonant (positive sample) and overall accuracy are used as metrics. Due to the imbalanced nature of training set, the CNN is trained with each sample weighted inversely proportional to class frequency. Relative error rate increment of performance over English has been calculated when models are applied on other languages. 

  
  \textbf{\emph{MFCCs}}: when calculating average MFCCs across the whole phonetic segment, MFCCs with dynamic coefficients (\emph{MC39}) achieved better accuracy and F1 score than MFCCs with only static coefficients (see first and third columns in Figs.~\ref{fig:sfig1} and \ref{fig:sfig2}). However, F1 and accuracy for MFCCs dropped by 5-20\% absolute when these models were evaluated on Spanish and Turkish. When calculating MFCCs anchored at landmark regions, \emph{MC39} obtained slightly better F1 score than its averaged model across the whole segment.

  
\textbf{\emph{Acoustic Cues vs.~MFCCs}}: in Figs.~\ref{fig:sfig1} and
\ref{fig:sfig2}, on English (training language), improvements are
clearly shown for acoustic cues compared with MFCC based
classifiers. When generalizing to other languages, acoustic cue based
features suffer much lower accuracy decrements than MFCCs. Furthermore, large error rate increase has been detected (up to 96.1\%) when MFCC-based model has been tested on other languages and the smallest error rate for it is 44.5\%. However, there is only up to 35.2\% relative error rate increase for Acoustic cues-based model.

\textbf{\emph{CNNs vs.~Feedforward}}: The filterbank used to compute
MFCCs can be viewed as a type of pre-determined convolutional network;
conversely, CNNs extract local patterns with trainable but fixed-length
convolutional windows. The last two columns in Figure
\ref{fig:sfig3} reveal that CNNs can hold
stable performance for each language, using 
either FFT or filterbank features as inputs. When applied to Spanish
and Turkish, CNNs show little drop in accuracy, while their F1 score
is higher in the test languages than in the training language;
the difference between overall accuracy and F1 score is apparently
an artifact of the highly non-uniform class distribution in Turkish
and Spanish, both of which have twice as many voiced as unvoiced
obstruents (see Table \ref{tab:corpus}).
  
\textbf{\emph{CNNs vs.~Acoustic Cues}}: Since the evaluation on two models (\emph{CNN+FFT} and \emph{CNN+FB}) results in similar accuracy scores, we consider \emph{CNN+FFT} as the best CNN model.
The last columns in Figs.~\ref{fig:sfig1}, ~\ref{fig:sfig2}, and ~\ref{tab:increase}
illustrate that the best CNN model outperforms acoustic cues across three
languages, in both metrics, and that it generalizes well to cross-lingual
corpora. 

\begin{table}[th]
    \caption{\label{tab:increase} {\it relative error rate increment (\%) on other languages.}}
    \vspace{-2mm}
    \centerline{
      \begin{tabular}{| c| c | c | }
        \hline
          &Turkish& Spanish   \\
        \hline 
      MFCC(13)(whole utterance)&  59.3& 54.8\\
      MFCC(13)(Landmark Region) & 40.9  &    44.5\\
      MFCC(39)(whole utterance)&  59.1& 96.4\\
      MFCC(39)(Landmark Region) & 50.0& 91.7\\
      Acoustic cues            &    31.3  &    35.2\\
      CNN                    &    16.2&    19.0\\
        \hline
      \end{tabular}
    }
    \vspace{-2mm}
  \end{table}

\section{CONCLUSION}
In this work, three different features are applied to build consonant voicing detectors, in order to test the theory that distinctive feature-based classes are robust over multilingual corpora.
MFCCs (in landmark region, and averaged over the whole phone utterance
duration), acoustic features extracted from the landmark region,
and features learned by a convolutional neural network (CNN) were tested
as features. Classifiers based on these features are all trained on English
and tested on English, Spanish, and Turkish.
Results show that MFCCs could not capture voicing
in either the training language or test languages.
Manual acoustic features generalize better to novel languages than MFCC.
Acoustic features learned by a CNN obtain best performance,
both on training languages and non-training languages.
We conclude that features capable of representing long-term spectral
dynamics relative to a phonetic landmark (CNN and landmark-based features)
are able to generalize cross-lingually with little or no loss of accuracy.

  \newpage
  \eightpt
  \bibliographystyle{IEEEtran}

  \bibliography{mybib}

\end{document}